\documentclass[dvipsnames,format=sigconf,anonymous=false,review=false]{acmart}
\usepackage{subcaption}

\renewcommand\footnotetextcopyrightpermission[1]{} 
\begin{document}
\pagestyle{plain} 

    \title{LLM-POET: Evolving Complex Environments using Large Language Models}

\author{Fuma Aki}
\orcid{0009-0008-8620-9695}

\author{Riku Ikeda}
\orcid{0009-0002-6919-3149}

\author{Takumi Saito}
\orcid{0009-0000-9838-0408}

\author{Ciaran Regan}
\orcid{0009-0001-8377-7231}

\author{Mizuki Oka}
\orcid{0000-0003-3915-4712}

\authornotemark[1]
\email{mizuki@cs.tsukuba.ac.jp}
\affiliation{%
  \institution{University of Tsukuba}
  \streetaddress{1 Chome-1-1 Tennodai}
  \city{Tsukuba}
  \state{Ibaraki}
  \country{Japan}
  \postcode{305-8577}
}

\renewcommand{\shortauthors}{Aki, et al.}

\begin{abstract}
Creating systems capable of generating virtually infinite variations of complex and novel behaviour without predetermined goals or limits is a major challenge in the field of AI. This challenge has been addressed through the development of several open-ended algorithms that can continuously generate new and diverse behaviours, such as the POET and Enhanced-POET algorithms for co-evolving environments and agent behaviour. One of the challenges with existing methods however, is that they struggle to continuously generate complex environments. In this work, we propose LLM-POET, a modification of the POET algorithm where the environment is both created and mutated using a Large Language Model (LLM). By fine-tuning a LLM with text representations of Evolution Gym environments and captions that describe the environment, we were able to generate complex and diverse environments using natural language. We found that not only could the LLM produce a diverse range of environments, but compared to the CPPNs used in Enhanced-POET for environment generation, the LLM allowed for a 34\% increase in the performance gain of co-evolution. This increased performance suggests that the agents were able to learn a more diverse set of skills by training on more complex environments.
\end{abstract}



\maketitle

\section{Introduction}
Recent advancements in AI have led to the development of agents that can succeed in challenging, yet domain-specific tasks. Once the domain has been mastered, however, learning typically ends. This domain-specific learning is in contrast to the real world, which is an ever-changing, open-ended ecosystem of new problems which require novel solutions. This challenge has been addressed through the development of several open-ended algorithms that can continuously generate new and diverse behaviours~\cite{NoveltySearch,NSLC2011,MapElites2015}. Among the various open-ended evolution (OEE) algorithms developed, co-evolution, the simultaneous evolution of the agent and the environment, is a remarkably advantageous approach, as demonstrated in studies such as hide and seek~\cite{baker2019emergent} or minimal criterion co-evolution~\cite{brant2017minimal,brant2020diversity}. 

To evaluate and compare co-evolution algorithms, Evolution Gym was developed, which offers a benchmark for co-optimizing the design and control of soft-robot agents~\cite{Evogym}. In this benchmark, both the agent and the environment are composed of different types of voxels (soft, rigid, actuators, empty), resulting in a vast space of possible agent and environment configurations.

Paired Open-Ended Trailblazer (POET) \cite{POET} and its successor Enhanced-POET \cite{wang2020enhanced} are co-evolutionary algorithms originally introduced to solve challenges in OpenAI Gym environments, such as the ``Bipedal Walker'' environment, but more recently have been implemented in Evolution Gym~\cite{oka2023beginning}. In these algorithms, both the agent and the environment co-evolve, with agent-environment pairs optimised before agents are transferred to other environments in the population for further learning.  One of the challenges with these POET algorithms, however, is the difficulty in continually generating complex environments, as the level of environmental difficulty tends to reach an upper limit.

To address this limitation, we propose LLM-POET, a modification of the Enhanced-POET algorithm in which the environments are created and mutated using a LLM instead of a CPPN. By fine-tuning a LLM with environment-caption pairs, complex and varied environments can be generated using natural language. Not only can the LLM create a wide range of environments, our results suggest that implementing LLMs for environment generation in the POET algorithm leads to an increase in the effectiveness of co-evolution due to the diversity of environments created. 

\section{LLM-POET}
\label{sec:methods}
\begin{figure}
    \centering
  \includegraphics[width=.9\linewidth]{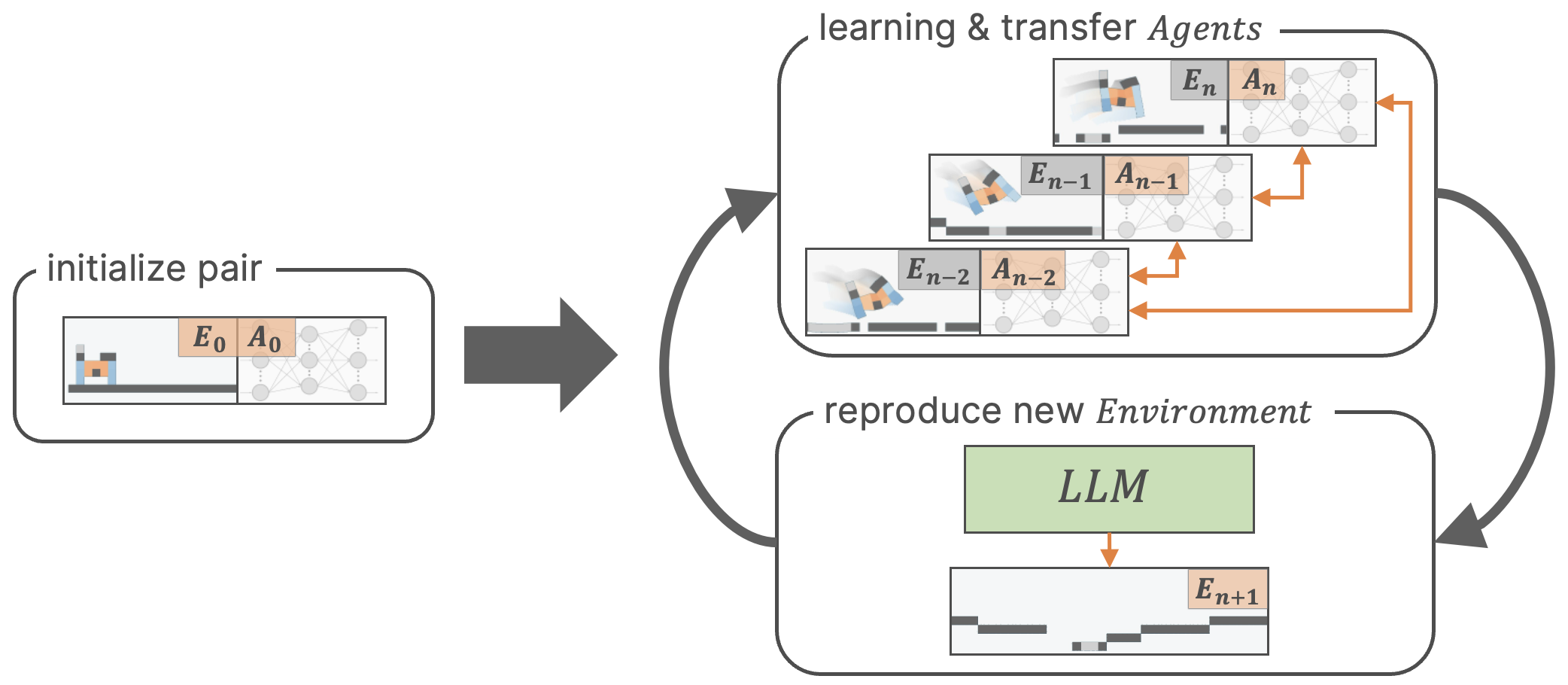}
  \caption{An overview of LLM-POET.}
  \Description{LLM-POET.}
  \label{fig:llm-poet}
\end{figure}

In this work, we propose LLM-POET, a modification of Enhanced-POET in which the environment generating CPPN is replaced with a LLM. The LLM takes as input a prompt that describes an environment in natural language and outputs a string representation of an Evolution Gym environment. In addition, we introduce an environment mutation method, making use of few-shot prompting and the LLM's ability to understand interesting environment mutations. These two features are then implemented into the POET algorithm, enabling the continuous evolution of complex environments through natural language. An overview of LLM-POET is depicted in Fig~\ref{fig:llm-poet}.

\subsection{Dataset Creation and LLM Fine-Tuning}
\label{sec:fine-tune}

\begin{figure}[ht]
    \centering
    \includegraphics[width=\linewidth]{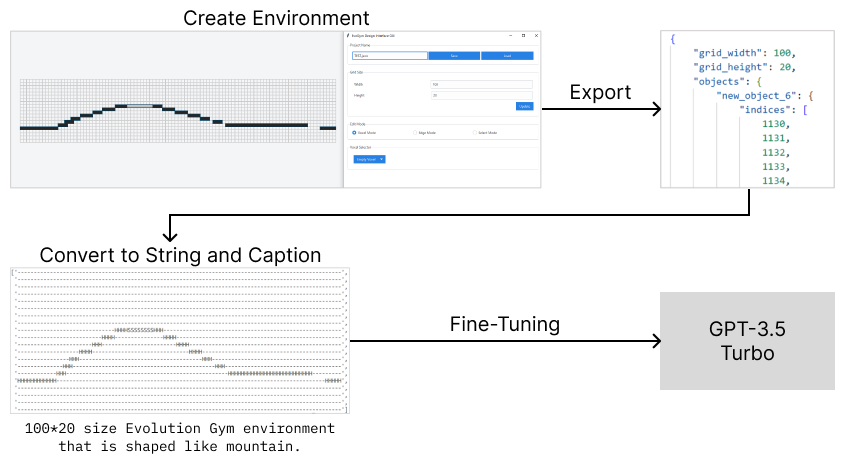}
    \caption{Overview of LLM fine-tuning.}
    \label{fig:fine-tune}
\end{figure}

To create a LLM capable of creating Evolution Gym environments, OpenAI's GPT-3.5 Turbo was fine-tuned using a dataset of environment-caption pairs. These environment caption-pairs were manually created using the Evolution Gym design tool GUI, and exported in JSON format. For each environment, a caption was also manually created, describing the shape and abstract difficulty of the environment, as well as its size. This process is illustrated in Fig~\ref{fig:fine-tune}. 

Taking inspiration from MarioGPT~\cite{sudhakaran2023mariogpt}, the proposed LLM takes as input the description of an environment and outputs a string containing characters representing the different types of voxels in Evolution Gym, such as `H' for rigid, `S' for soft or `-' for empty voxels. At inference time, instructions such as ``simple environment'', ``environment with many holes'', and `` difficult environment with mountains'' can be given, and the corresponding environment is output in two dimensions, as in Evolution Gym.

\begin{figure}[h]
    \centering
    \begin{subfigure}{.45\linewidth}
        \centering
        \includegraphics[width=\linewidth]{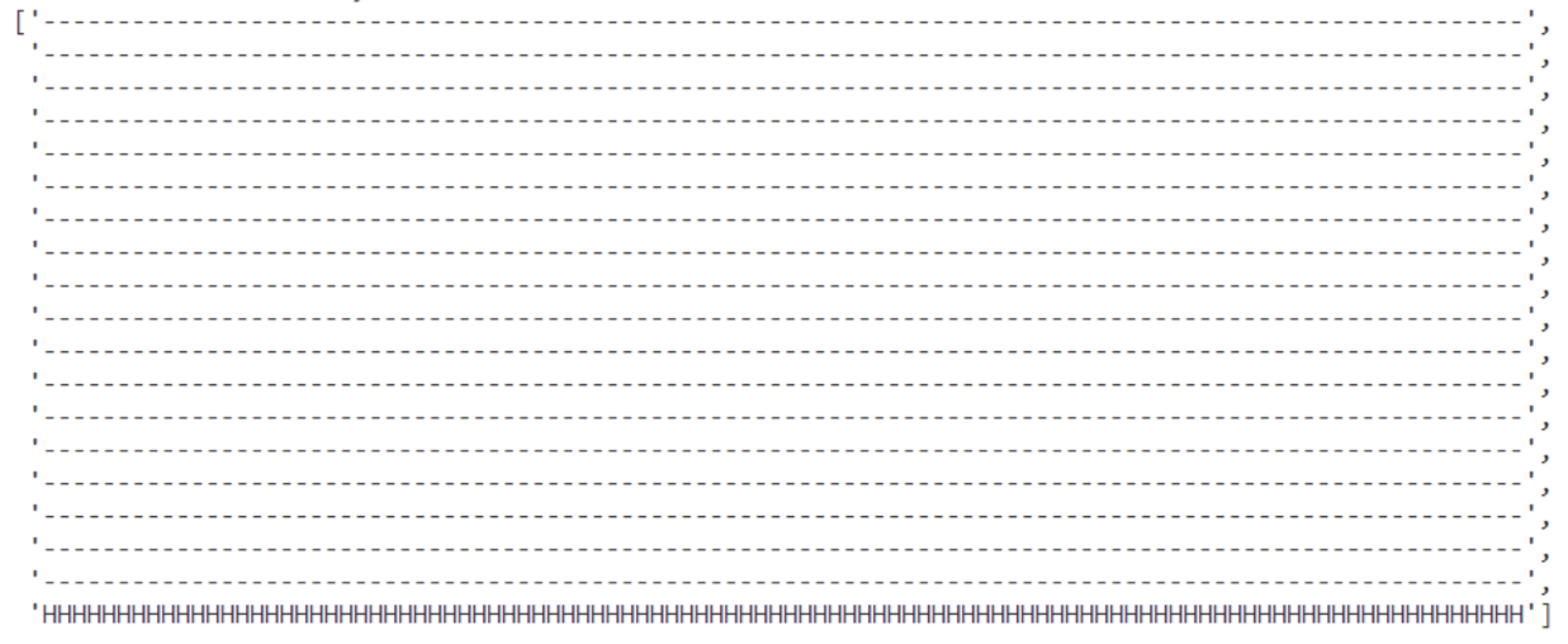}
        \caption{100*20 size Evolution Gym environment that consists of flat terrain.}
        \label{fig:sub1}
    \end{subfigure}
    \hspace*{2mm}
    \begin{subfigure}{.45\linewidth}
        \centering
        \includegraphics[width=\linewidth]{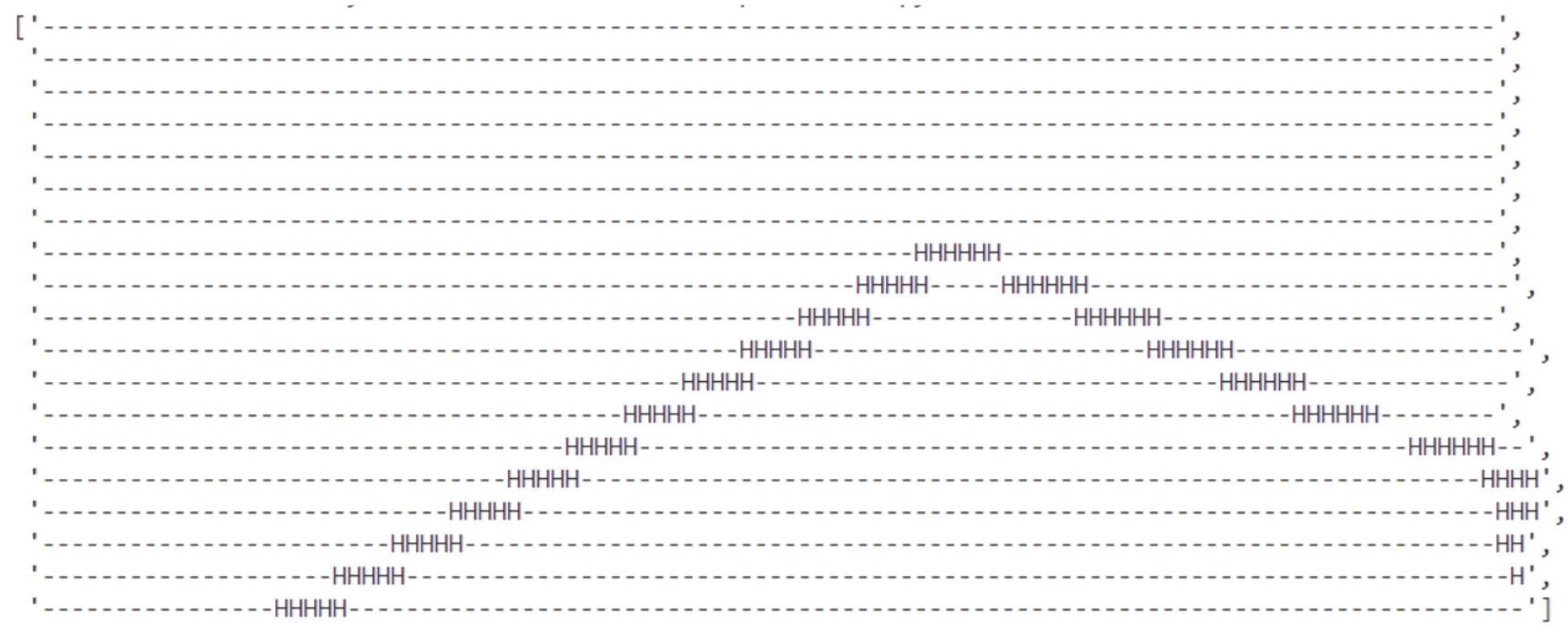}
        \caption{100*20 size Evolution Gym environment that is shaped like a pyramid.}
        \label{fig:sub2}
    \end{subfigure}
    
    \begin{subfigure}{.45\linewidth}
        \centering
        \includegraphics[width=\linewidth]{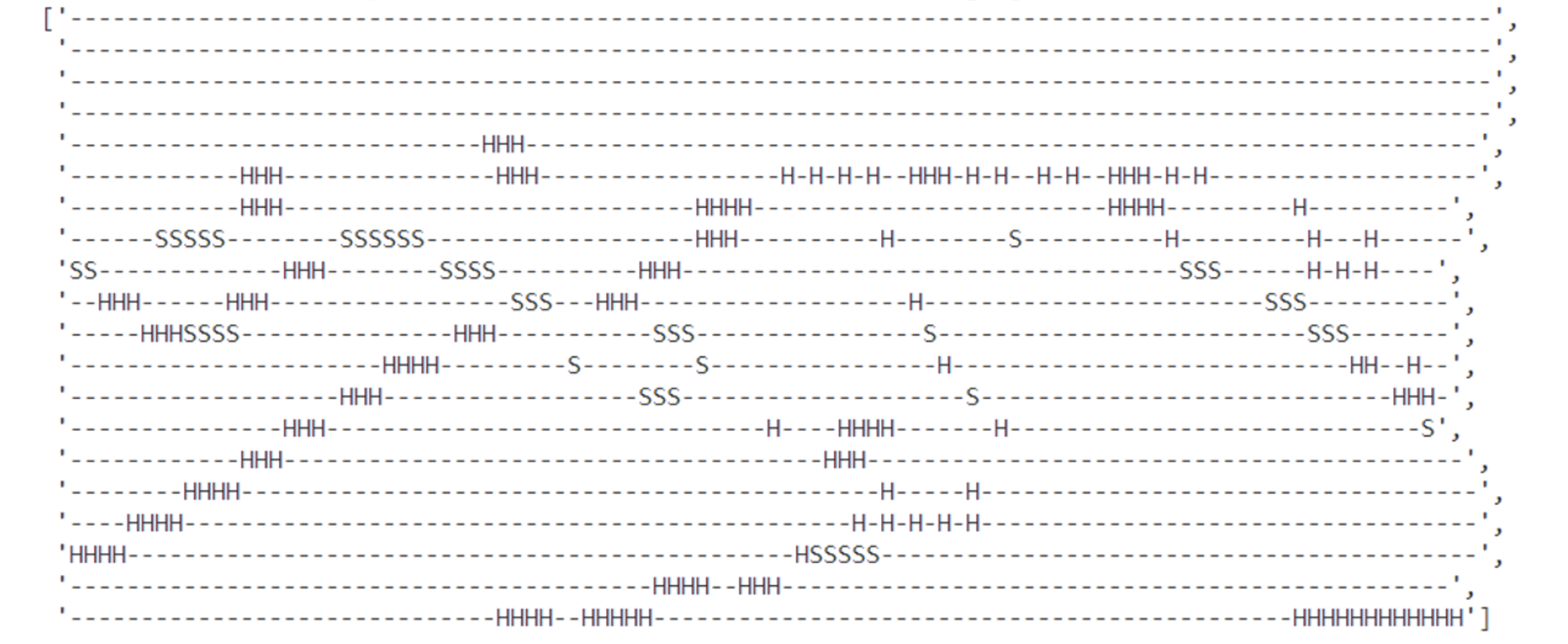}
        \caption{100*20 size Evolution Gym environment that is complex and challenging.}
        \label{fig:sub3}
    \end{subfigure}
    \hspace*{2mm}
    \begin{subfigure}{.45\linewidth}
        \centering
        \includegraphics[width=\linewidth]{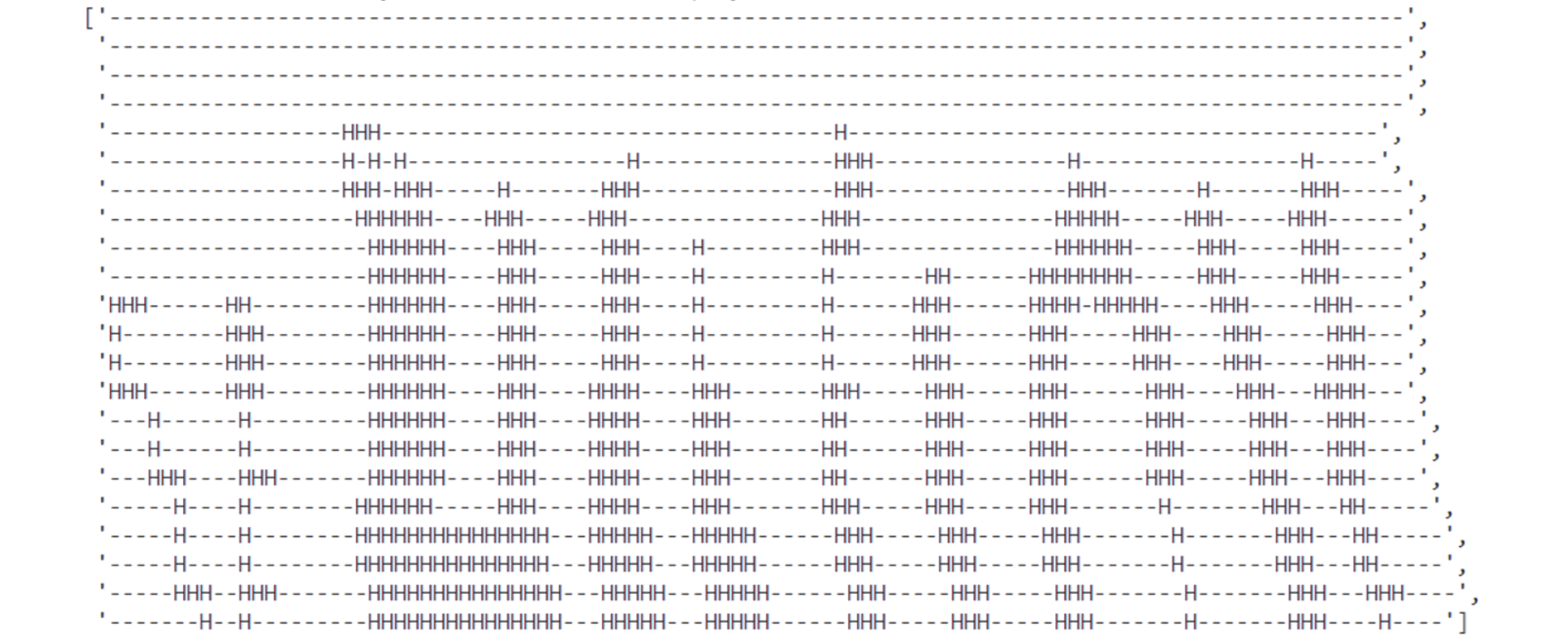}
        \caption{100*20 size Evolution Gym environment that is spiky.\phantom{xxxxxx}} 
        \label{fig:sub4}
    \end{subfigure}

    \caption{Environments generated using a variety of prompts with the fine-tuned LLM.}
    \label{fig:sample_envs}
\end{figure}

After fine-tuning however, the size of the environment output by the LLM does not always match the size requested in the prompt. Additionally, the LLM may occasionally output symbols which do not correspond to Evolution Gym voxels. To overcome these discrepancies, minor post-processing of the output string is necessary. After post-processing, the strings are converted into Evolution Gym environments, examples of which are shown in Fig~\ref{fig:sample_envs}. 

In addition to being able to create a wide range of environments, the fine-tuned LLM is shown to create environments with words such as ``pyramid'' or ``spiky'' without these words being used in the fine-tuning dataset. Furthermore, since there were no environments in the training dataset where the blocks were arranged vertically, vertical environments generated from words such as ``spiky'' in Fig~\ref{fig:sub4} demonstrate that fine-tuning allowed the LLM to understand the relationship between the caption and the environment.

\subsection{Integration with POET}
As POET is an open-ended learning algorithm which continuously alters existing environments, a method to evolve environments using a LLM is proposed. To achieve this environment mutation, two approaches are used. First, due to the nondeterministic nature of the LLM's output when a non-zero temperature is used, an environment can be mutated by simply prompting the LLM with the same initial prompt. Second, to increase the complexity of environments, mutating the input prompt itself is also considered. Mutating the prompt using LLMs is inspired by \cite{bradley2023qualitydiversity}, which showed that LLMs have demonstrated an ability to recognise interesting mutations. To achieve this mutation, few-shot prompting was used to alter the original prompt.

\begin{figure}[h]
    \centering
    \includegraphics[width=0.35\linewidth]{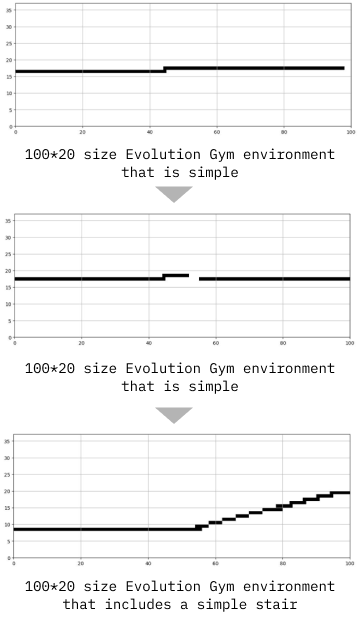}
    \caption{Mutating environments with the original or mutated prompts.}
    \label{fig:mutation}
\end{figure}

\begin{figure}[]
    \centering
    \includegraphics[width=.7\linewidth]{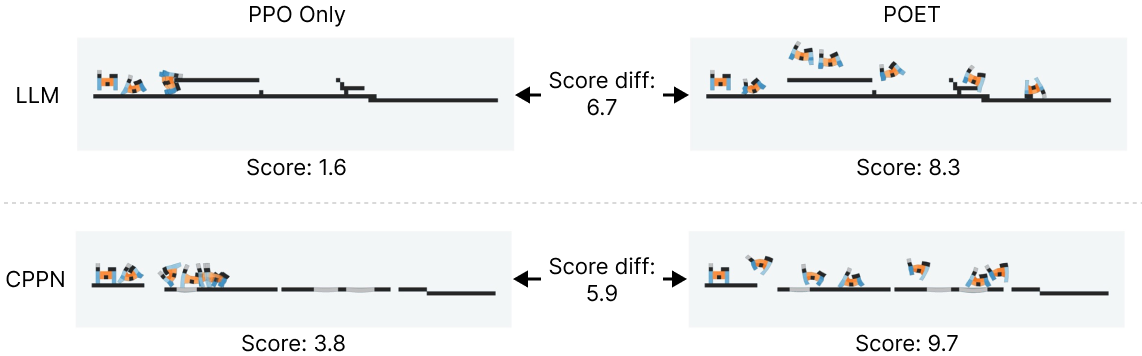}
    \caption{Calculating score differences. The difference between PPO only score (left) and the POET score (right) quantifies the ability of the environment generation model to create diverse environments, leading to improved agent learning.}
    \label{fig:evaluation}
\end{figure}

When integrating this mutation into the Enhanced-POET algorithm, either the original prompt or a mutated prompt will be input to the LLM with a 50\% probability. An example of this prompt and environment mutation is shown in Fig~\ref{fig:mutation}. The environment-generating LLM and the environment mutation method can then be implemented in the POET algorithm to allow for continually evolving environments. 

\section{Experiment}
\label{sec:experiment}
To evaluate LLM-POET and Enhanced-POET, direct comparison of scores is impractical as the POET algorithm is continuously generating different environments. To overcome this, the following evaluation method is proposed. For each environment generated by Enhanced-POET and LLM-POET, an agent is trained using PPO alone, without co-evolution. By comparing the best score of the agent trained with PPO only, to the score of the agent trained with either Enhanced-POET or LLM-POET, the performance gain of co-evolution in that environment can be measured. This is depicted in Fig~\ref{fig:evaluation}.

In this evaluation, the difference in scores obtained by the agents trained using the POET algorithm and the PPO algorithm can be considered as the effect of learning from the POET algorithm. This is because the PPO algorithm is used when the agent learns the environment within the POET algorithm, and the difference in scores can be attributed to the effect of the agent learning while transferring between environments. Therefore, in this experiment, where the only difference is the environment generation method, it can be said that the method with more diverse environments had a larger difference between the POET and PPO scores.

The experiment and evaluation procedure can then be summarized as:

\begin{enumerate}
    \item Use LLM to create an environment for use in the POET algorithm (LLM-POET).
    \item Obtain the score of the agent in the environment using the POET algorithm.
    \item Extract each environment, perform reinforcement learning with PPO five times, and compare the best score with that of POET.
    \item Repeat the above with the CPPNs for environment generation (Enhanced-POET).
    \item Compare environment generation methods by comparing the score differences of step 3.
\end{enumerate}

with the POET algorithm taking the following form:

\begin{enumerate}
    \item  Generate 10 initial environment pairs.
    \item Perform 30 steps of PPO learning for all environment-agent pairs. PPO performs learning four times in each iteration. The score of an agent in an environment is determined by how much distance the agent has travelled along the x-axis.
    \item Transfer each agent to another environment. If the agent in the new environment has a better score than the agent originally trained in the environment, replace the agent.
    \item Randomly select a single high-scoring environment and generate a new environment using either CPPNs or a LLM.
\end{enumerate}

\noindent Steps 2-4 were repeated 100 times to train agents on a vast range of environments.

\section{Results}

\begin{figure*}[]
    \centering
    \begin{subfigure}[b]{\linewidth}
        \centering
        \includegraphics[width=.6\linewidth]{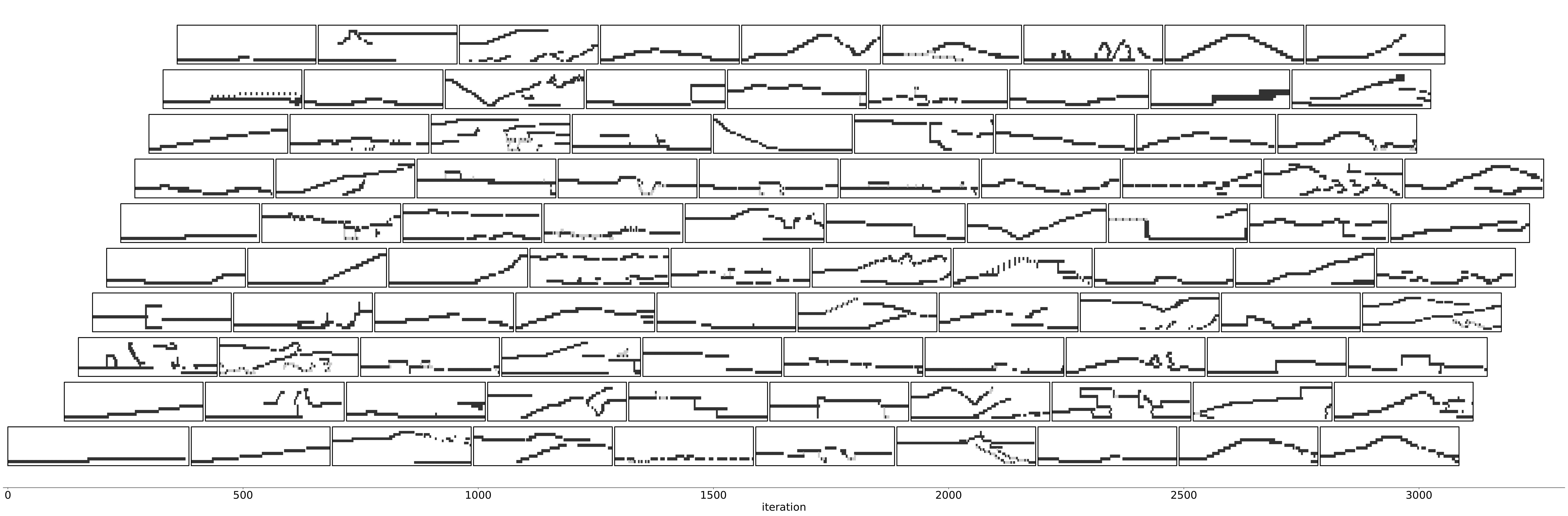}
        \caption{All environments generated using LLM-POET.}
        \label{fig:niche_llm}
    \end{subfigure}
    \vspace{1em} 

    \begin{subfigure}[b]{\linewidth}
        \centering
        \includegraphics[width=.6\linewidth]{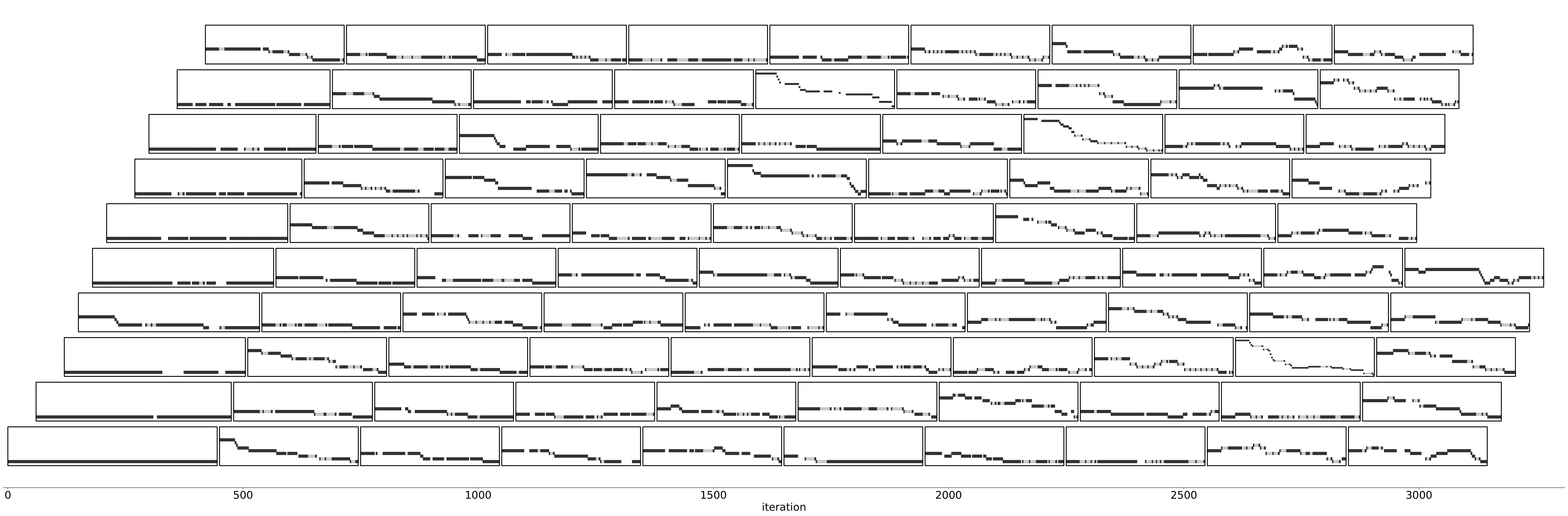}
        \caption{All environments generated using Enhanced-POET.}
        \label{fig:niche_cppn}
    \end{subfigure}
    \caption{Comparison of environments generated by LLM-POET and Enahnced-POET}
\end{figure*}

The LLM-POET algorithm resulted in 97 unique environments, with the Enhanced-POET algorithm resulting in 95 environments, shown in niche history in Fig~\ref{fig:niche_llm} and ~\ref{fig:niche_cppn} respectively. The difference between the best PPO scores and the best POET scores for these environments was then calculated. The average score difference was found to be $1.11 \pm 0.42$ for LLM-POET, and $0.83 \pm 0.42$ for Enhanced-POET, a 34\% increase in the performance gain of co-evolution when using LLMs for environment generation. One possible explanation for this is the increased complexity of environment in the niche history.

\begin{figure}[t]
    \centering
    \includegraphics[width=0.8\linewidth]{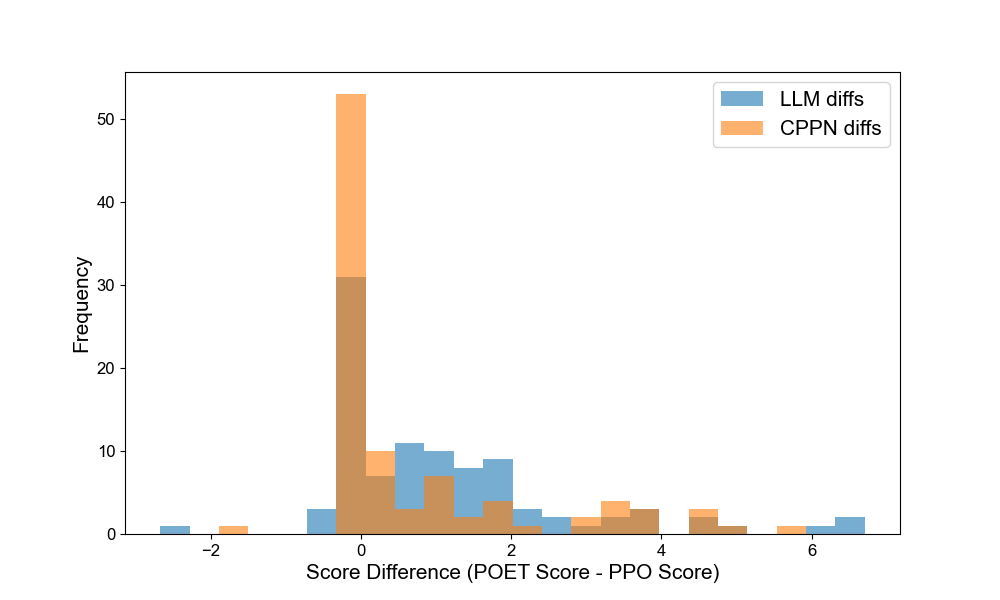}
    \caption{The distribution of score differences.}
    \label{fig:distribution}
\end{figure}

Additionally, the histogram of score differences is shown in Fig~\ref{fig:distribution}. The skewness of these distributions was found to be $1.517$ and $1.652$ for the LLM-POET and Enhanced-POET difference respectively, with the lower value for LLM-POET indicating that it is slightly more left-skewed with a greater difference in scores. Although the uncertainty in the result is high, the 34\% increase in the POET performance gain suggests that the agents were able to train on more complex environments using LLMs for environment generation.


\section{Conclusion}
In this work, we introduce LLM-POET, a modification of the Enhan\-ced-POET algorithm, in which a LLM is used for both environment generation and mutation. Not only have we shown that a fine-tuned model can generate Evolution Gym compatible environments, but when implemented into an open-ended learning algorithm such as POET, this environment generation method can lead to an increase in the effectiveness of open-ended learning.

To evaluate the effectiveness of LLMs for open-ended learning, LLM-POET is compared to Enhanced-POET, which uses CPPNs for environment generation. Our results suggest that LLM-POET has a 34.4\% increase in the effectiveness of open-ended learning when compared to Enhanced-POET. This improvement in the performance of the POET algorithm indicates that the environments generated by LLM-POET have more diversity, which allows the agents to learn more sophisticated behaviors, addressing one of the key challenges in open-ended research.

\begin{acks}
A special thanks to Nanami Iwahashi for assistance with the figures, enhancing the clarity of this work.
\end{acks}

\bibliographystyle{plain}
\bibliography{sample-base}


\end{document}